\documentclass[letterpaper]{article}
\usepackage[preprint]{aaai2027}
\usepackage[hyphens]{url}
\usepackage{graphicx}
\usepackage{natbib}
\usepackage{caption}
\usepackage{amsmath}
\usepackage{amssymb}
\usepackage{multirow}
\title{Tri-DehazeGS: Scene--Medium Decoupled Gaussian Splatting with Transmittance-Aware Optimization}
\author{
Kui Jiang,\textsuperscript{\rm 1}
Yang Gu,\textsuperscript{\rm 1}
Jiacheng Liu,\textsuperscript{\rm 1}
Shiyu Liu,\textsuperscript{\rm 1}
Youyu Chen,\textsuperscript{\rm 1}
Hui Liu\textsuperscript{\rm 2}
}
\affiliations{
\textsuperscript{\rm 1}Harbin Institute of Technology\\
\textsuperscript{\rm 2}Meituan
}

\newcommand{\qualmethodlabels}{%
{\footnotesize
\makebox[0.98\textwidth][c]{%
\makebox[0.14\textwidth][c]{3DGS}%
\makebox[0.14\textwidth][c]{BiLaLoRA-GS}%
\makebox[0.14\textwidth][c]{I$^2$-NeRF}%
\makebox[0.14\textwidth][c]{Plenodium}%
\makebox[0.14\textwidth][c]{3D-UIR}%
\makebox[0.14\textwidth][c]{Ours}%
\makebox[0.14\textwidth][c]{GT}%
}}}
\begin{document}
\maketitle

\begin{abstract}

Recovering clean 3D scenes from hazy multi-view images is challenging because haze attenuates scene radiance and introduces atmospheric scattering. Recent scattering-aware Gaussian Splatting methods introduce physical haze models into reconstruction, but they often apply degradation in image space or bind medium-related variables to Gaussian primitives, which can entangle clean scene radiance with atmospheric effects. Moreover, low-transmittance regions provide weakened supervision for Gaussian optimization, causing distant or dense-haze areas to be under-reconstructed. We argue that clean reconstruction under haze requires both scene--medium disentanglement and transmittance-aware optimization rebalancing. To this end, we propose Tri-DehazeGS, a scene--medium decoupled Gaussian Splatting framework. It represents the clean scene with Gaussian primitives, models the participating medium using an independent view-shared tri-plane field, and composes hazy observations through a physical scattering model. We further introduce Medium-Decoupled Transmittance Gradient Compensation (MD-TGC), which compensates haze-suppressed gradients after medium freezing without altering forward rendering. Experiments on real and synthetic haze benchmarks show that Tri-DehazeGS improves clean novel-view reconstruction. Code is available at \url{https://github.com/aptx46/Tri-DehazeGS}.

\end{abstract}

\begin{figure}[t]
\centering
\includegraphics[width=\columnwidth]{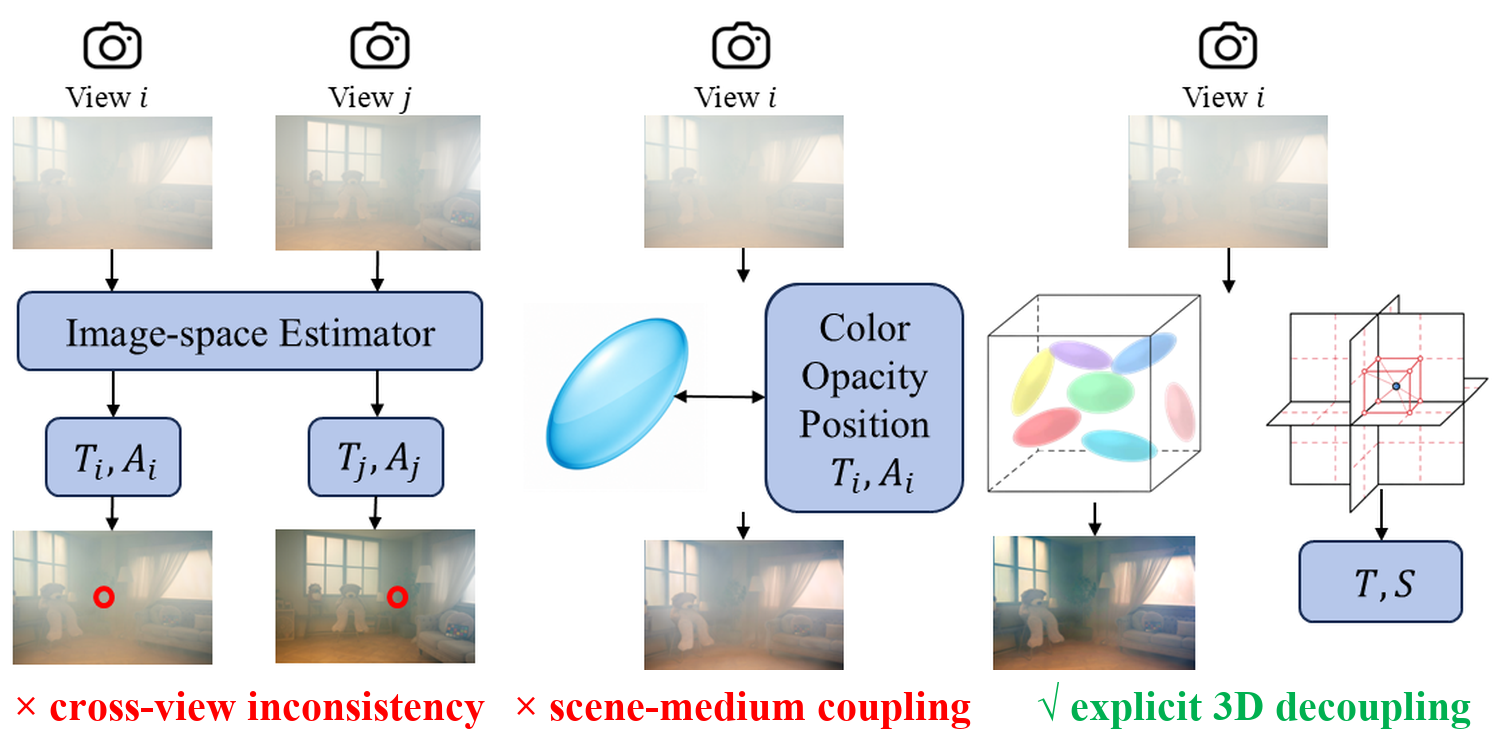}\\[-1pt]
\vspace{-1mm}
{\footnotesize
\makebox[\columnwidth][c]{%
\makebox[0.333\columnwidth][c]{(a) Image-space}%
\makebox[0.333\columnwidth][c]{(b) Gaussian-bound}%
\makebox[0.333\columnwidth][c]{(c) Ours}%
}}
\caption{Medium parameterization paradigms. Image-space estimators predict degradation variables per view and lack an explicit shared 3D medium representation; Gaussian-bound methods attach medium variables to Gaussian primitives; our framework separates the clean Gaussian scene from the tri-plane medium field and composes hazy observations through the atmospheric scattering model.}
\label{fig:motivation_teaser}
\vspace{-4mm}
\end{figure}

\section{Introduction}

Real-world multi-view reconstruction is often performed in open environments, where haze alters image formation through transmittance attenuation and atmospheric scattering~\cite{narasimhan2002vision}. These effects degrade distant structures, scene contrast, and color appearance, motivating classical and learning-based image dehazing studies~\cite{he2011darkchannel,li2017aodnet}. Therefore, hazy scene reconstruction should not merely reproduce degraded observations; it should recover clean novel-view renderings and a haze-aware Gaussian representation while explicitly accounting for the participating medium.

Neural rendering has greatly advanced multi-view scene reconstruction~\cite{mildenhall2020nerf,yu2022plenoxels}. NeRF-based approaches can model participating media through implicit fields and volumetric integration~\cite{ramazzina2023scatternerf,teigen2023fognerf}, but dense ray sampling and repeated network queries lead to high training and rendering costs. In contrast, 3D Gaussian Splatting (3DGS) uses explicit primitives, differentiable rasterization, and direct Gaussian-level optimization to achieve high-quality real-time novel-view synthesis~\cite{kerbl2023gaussian}, making it an attractive backbone for practical reconstruction under challenging weather.

However, clean reconstruction under haze differs from conventional novel-view synthesis. Clean inputs can be approximated as direct observations of scene radiance, whereas hazy images are jointly generated by the underlying scene and the participating medium. If standard 3DGS is optimized directly on hazy images, atmospheric attenuation and scattering may be absorbed into Gaussian colors, opacities, scales, or geometry. Such a representation may reproduce hazy views, but it can struggle to recover the clean scene hidden behind the medium.

We argue that haze affects Gaussian reconstruction at both the representation and optimization levels. At the representation level, scene appearance and atmospheric scattering may be encoded into Gaussian primitives simultaneously, producing an entangled representation~\cite{jain2026smokeseer,xu2026dehazesplat}. At the optimization level, low transmittance attenuates the gradients propagated to the clean Gaussian branch, biasing appearance and opacity updates toward nearby clear regions while leaving distant and heavily degraded structures under-optimized.

Existing methods only partially address these challenges. 
The first applies haze degradation in image space after rendering clean views, which lacks a consistent 3D medium representation across views (Figure~\ref{fig:motivation_teaser} a). The second incorporates medium-related variables directly into Gaussian primitives through per-primitive medium attributes or auxiliary transformation networks ~\cite{yu2025dehazegs,li2026nimbusgs}, which couples scene content and medium effects within the same primitives (Figure~\ref{fig:motivation_teaser} b). In both cases, the fidelity of the recovered clean scene depends on how well scene and medium are separated during joint optimization. 

These observations suggest that hazy Gaussian reconstruction should be viewed as a joint problem of \textbf{scene--medium disentanglement} and \textbf{supervision rebalancing}. The former requires learning an explicit three-dimensional participating-medium representation that is independent of Gaussian scene primitives, while the latter requires restoring the optimization signals suppressed by transmittance attenuation. Only by addressing both aspects simultaneously can a clean and physically meaningful scene representation be reliably recovered.

To this end, we propose \textbf{Tri-DehazeGS}, a unified framework for clean novel-view reconstruction from hazy multi-view images.Tri-DehazeGS jointly learns a decoupled representation of the participating medium and optimizes the clean scene with a transmittance-aware strategy. Specifically, we introduce a compact tri-plane~\cite{chan2022eg3d} fog field that independently represents spatial extinction and direction-aware medium radiance within a shared three-dimensional coordinate system(Figure~\ref{fig:motivation_teaser} c). The learned medium representation provides a physically motivated explanation of haze across views while discouraging atmospheric effects from being absorbed into Gaussian primitives. Furthermore, we propose \textbf{Medium-Decoupled Transmittance Gradient Compensation (MD-TGC)}, which freezes the learned medium representation after medium stabilization and compensates the gradients attenuated in low-transmittance regions during backpropagation without altering the forward image formation process. By jointly addressing representation disentanglement and optimization rebalancing, Tri-DehazeGS improves clean appearance recovery and novel-view rendering quality.

The main contributions of this work are summarized as
\begin{itemize}
    \item We formulate hazy 3D Gaussian reconstruction as a joint problem of \textbf{scene--medium disentanglement} and \textbf{supervision rebalancing}, providing a structured perspective for clean novel-view reconstruction under participating media.

    \item We propose \textbf{Tri-DehazeGS}, a unified Gaussian reconstruction framework that jointly learns an explicit view-shared participating-medium representation and clean Gaussian scene primitives. A compact tri-plane fog field organizes haze variables in a shared 3D space and reduces the entanglement between atmospheric degradation and scene representation.
    
    \item We develop \textbf{Medium-Decoupled Transmittance Gradient Compensation (MD-TGC)}, a transmittance-aware optimization strategy that compensates attenuated gradients after medium freezing without modifying the physically based forward rendering process.
\end{itemize}

\section{Method}
\paragraph{Problem Formulation.}
Given calibrated hazy views
$\mathcal{D}=\{(\mathbf{I}_i,\Pi_i)\}_{i=1}^{N}$,
where $\Pi_i$ denotes camera parameters and $N$ is the number of training views, Tri-DehazeGS aims to recover a clean Gaussian scene
$\mathcal{G}$ and a view-shared participating medium field
$\mathcal{M}_{\theta}$ using only hazy supervision.

The key idea is to address two issues in scattering-aware reconstruction:
(1) \textbf{scene--medium entanglement}, where atmospheric effects may be
implicitly absorbed into Gaussian primitives during joint reconstruction, and
(2) \textbf{transmittance-induced supervision attenuation}, where the scattering
composition weakens optimization signals for distant and dense-haze regions.

Tri-DehazeGS separates scene and medium using two branches. For a pixel $\mathbf{p}$, the Gaussian branch renders clean radiance $\hat{\mathbf{J}}(\mathbf{p})$, while the medium field predicts transmittance $\hat{T}(\mathbf{p})$ and scattering residual
$\hat{\mathbf{S}}(\mathbf{p})$:

\begin{equation}
\hat{\mathbf{I}}(\mathbf{p})
=
\hat{\mathbf{J}}(\mathbf{p})\hat{T}(\mathbf{p})
+
\hat{\mathbf{S}}(\mathbf{p}).
\label{eq:haze_render}
\end{equation}

The clean scene is obtained by rendering $\hat{\mathbf{J}}$ without the
medium branch, while $\hat{T}$ is further used as a visibility-aware
optimization signal.

\begin{figure*}[t]
\centering
\includegraphics[width=0.94\textwidth]{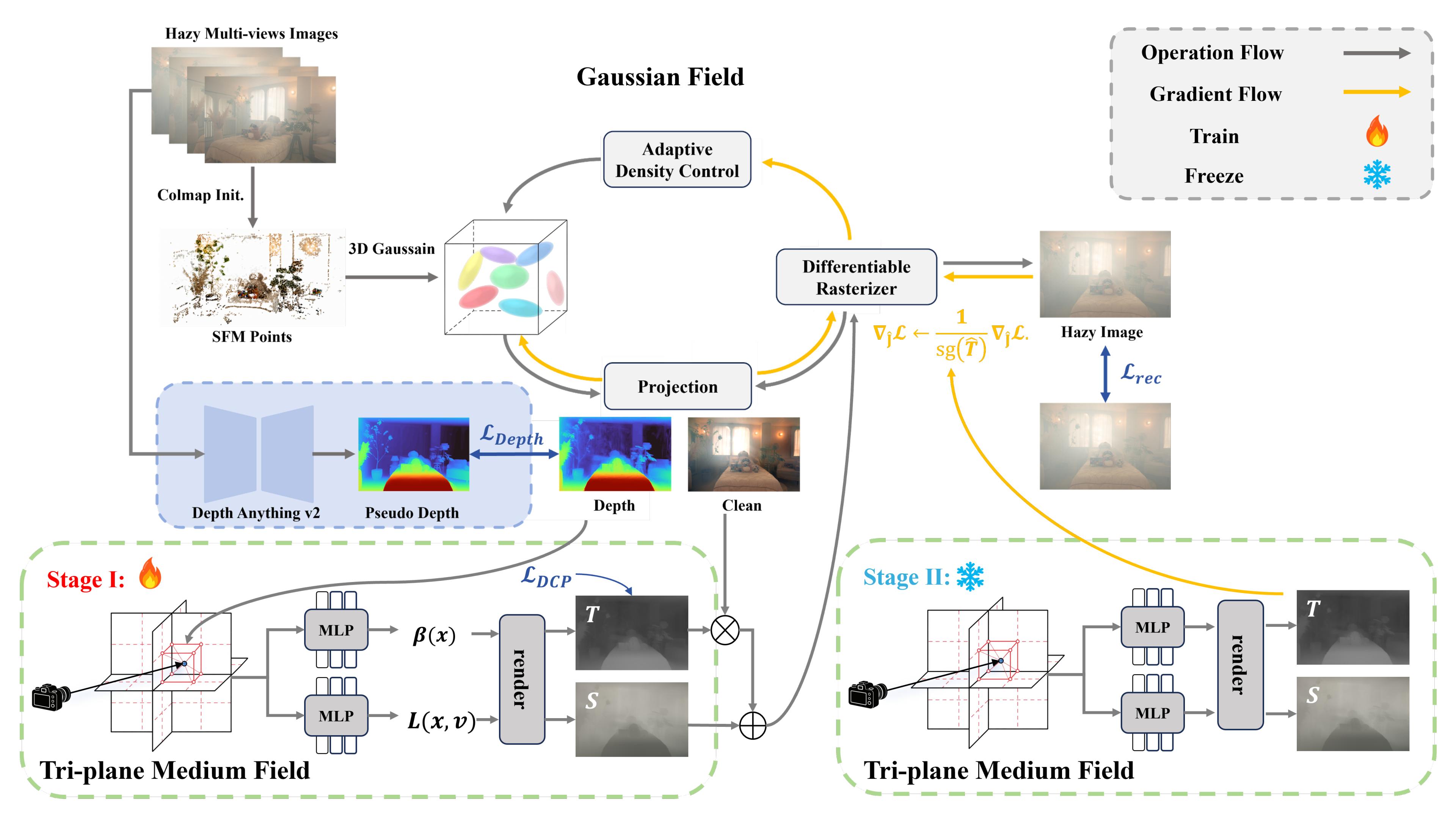}\vspace{-2mm}
\caption{
\textbf{Overview of Tri-DehazeGS.}
The framework jointly addresses scene--medium entanglement and transmittance-imbalanced supervision.
\textbf{Stage I} learns a decoupled Gaussian scene and view-shared medium field through physical haze rendering.
\textbf{Stage II} freezes the medium field and applies MD-TGC to compensate transmittance-suppressed gradients without changing forward image formation.
}\vspace{-4mm}
\label{fig:overview_pipeline}
\end{figure*}

\subsection{Scene--Medium Decoupled Gaussian Rendering}
3D Gaussian Splatting represents a scene using explicit Gaussian primitives:
\begin{equation}
\mathcal{G}
=
\{G_i=(\boldsymbol{\mu}_i,
\boldsymbol{\Sigma}_i,
\alpha_i,
\mathbf{c}_i)\}_{i=1}^{M}.
\end{equation}

The clean radiance is rendered through differentiable alpha compositing:
\begin{equation}
\hat{\mathbf{J}}(\mathbf{p})
=
\sum_i
\tau_i^G
\alpha_i
\mathbf{c}_i(\mathbf{v}),
\quad
\tau_i^G=
\prod_{j<i}(1-\alpha_j).
\label{eq:gaussian_render}
\end{equation}

However, directly optimizing hazy observations with Eq.~\eqref{eq:gaussian_render} allows haze effects to be encoded into Gaussian appearance and geometry. To avoid this ambiguity, Tri-DehazeGS assigns different explanatory roles: the Gaussian branch models intrinsic scene radiance, while the medium branch models attenuation and in-scattering.

\subsection{View-Shared Tri-Plane Medium Field}
Real haze exhibits spatially varying density and view-dependent illumination.
We therefore represent the participating medium using a continuous
three-dimensional field.

For a sampled point $\mathbf{x}$ along a camera ray, the field predicts:
(1) extinction coefficient $\beta(\mathbf{x})$, and
(2) directional medium radiance $\mathbf{L}(\mathbf{x},\mathbf{v})$.
We adopt a tri-plane representation to learn the corresponding medium feature:
\begin{equation}
\mathbf{f}(\mathbf{x})
=
\operatorname{Concat}
\left[
\mathbf{f}_{xy}(\mathbf{x}),
\mathbf{f}_{xz}(\mathbf{x}),
\mathbf{f}_{yz}(\mathbf{x})
\right],
\label{eq:triplane_feature}
\end{equation}
where $\operatorname{Concat}$ denotes channel-wise concatenation, $\mathbf{f}_{ab}(\mathbf{x})=\mathcal{F}_{ab}(\pi_{ab}(\mathbf{x}))$ for $ab\in\{xy,xz,yz\}$, $\mathcal{F}_{ab}$ is the learnable feature plane, and $\pi_{ab}$ denotes projection onto the corresponding plane coordinates.

Unlike image-space degradation prediction, the tri-plane field places all observations in a common three-dimensional medium coordinate system.
Consequently, different views query the same atmospheric structure, encouraging cross-view consistency while retaining the efficiency of a compact explicit representation.

The extinction head $h_{\beta}$ maps the tri-plane feature to a non-negative extinction coefficient:
\begin{equation}
\beta(\mathbf{x})
=
\kappa\,
\operatorname{softplus}
\left(h_{\beta}(\mathbf{f}(\mathbf{x}))\right),
\label{eq:medium_heads}
\end{equation}
where
$\operatorname{softplus}$ enforces non-negativity, and $\kappa>0$ is a learnable global extinction scale. This separates the relative spatial distribution of haze from its global strength.

The medium-light head $h_L$ predicts direction-aware radiance with low-order SH:
\begin{equation}
\mathbf{L}(\mathbf{x},\mathbf{v})
=
\operatorname{sigmoid}
\left(
\operatorname{SH}
\bigl(h_L(\mathbf{f}(\mathbf{x})),\mathbf{v}\bigr)
\right),
\label{eq:medium_light}
\end{equation}
where
$SH$ denotes low-order spherical-harmonic evaluation, and $\operatorname{sigmoid}$ is a range-limiting activation for RGB radiance.

For ray samples $\{\mathbf{x}_k\}_{k=1}^{K}$ with interval length
$\Delta_k$, the accumulated transmittance is
\begin{equation}
\hat{T}
=
\exp
\left(
-\sum_{k=1}^{K}\beta_k\Delta_k
\right).
\label{eq:ray_transmittance}
\end{equation}
The in-scattered residual is accumulated as
\begin{equation}
\hat{\mathbf{S}}
=
\sum_{k=1}^{K}
T_k
\left(1-\exp(-\beta_k\Delta_k)\right)
\mathbf{L}_k,
\label{eq:scattering_residual}
\end{equation}
where $\hat{T}$ is the final accumulated transmittance, $T_k$ is the attenuation before the $k$-th interval, and \(1-\exp(-\beta_k\Delta_k)\) is the interval scattering opacity.
Equations above provide a physically consistent decomposition of clean
radiance, attenuation, and in-scattering.

\subsection{Medium-Decoupled TGC}
Although scene--medium decomposition resolves representation ambiguity,
haze introduces another optimization problem: low-transmittance regions receive weaker Gaussian supervision~\cite{wang2025uw}.

From Eq.~\eqref{eq:haze_render},
\begin{equation}
\frac{\partial\mathcal{L}}
{\partial\hat{\mathbf{J}}(\mathbf{p})}
=
\hat{T}(\mathbf{p})
\frac{\partial\mathcal{L}}
{\partial\hat{\mathbf{I}}(\mathbf{p})},
\label{eq:gradient_attenuation}
\end{equation}
where $\mathcal{L}$ is the training objective and $\mathbf{p}$ denotes an image pixel.
Therefore, transmittance acts as a gradient attenuation factor. Distant structures and dense-haze regions may have large reconstruction errors but receive insufficient updates due to small $\hat{T}$.

This observation reveals a second role of transmittance.
Beyond describing image formation, it quantifies the strength of supervision reaching the clean scene branch.
Unlike depth-only weighting, learned transmittance jointly reflects long propagation distance and locally dense participating media.
We therefore introduce \textbf{Medium-Decoupled Transmittance Gradient Compensation} (MD-TGC), which rescales only the backward gradient of the clean Gaussian rendering:

\begin{equation}
\nabla_{\hat{\mathbf{J}}}\mathcal{L}
\leftarrow
\frac{1}
{\operatorname{clamp}
(\hat{T},T_{\min})}
\nabla_{\hat{\mathbf{J}}}\mathcal{L},
\label{eq:md_tgc}
\end{equation}
where $\nabla_{\hat{\mathbf{J}}}\mathcal{L}$ is the gradient entering the clean Gaussian rendering and $\operatorname{clamp}(\cdot,T_{\min})$ prevents a near-zero denominator.

MD-TGC only modifies backward propagation and keeps haze rendering in Eq.~\eqref{eq:haze_render} unchanged. Thus, it preserves the physical rendering model while compensating the supervision attenuation caused by haze. Importantly, the medium field is frozen before applying MD-TGC to prevent degenerate solutions where the medium artificially reduces transmittance to amplify Gaussian gradients. After freezing $\mathcal{M}_{\theta}$, transmittance becomes a fixed and medium-decoupled visibility signal for clean-scene refinement.

\subsection{Two-Stage Optimization}
Tri-DehazeGS follows a two-stage schedule aligned with its two objectives: first establishing a physically motivated scene--medium decomposition, and then rebalancing clean-scene supervision.

\paragraph{Stage I: scene--medium decomposition.}
The Gaussian scene and medium field are jointly optimized by reconstructing the input hazy views:
\begin{equation}
\mathcal{L}_{\mathrm{rec}}
=
(1-\lambda_{\mathrm{ssim}})
\|\hat{\mathbf{I}}-\mathbf{I}\|_1
+
\lambda_{\mathrm{ssim}}
\bigl(1-\operatorname{SSIM}
(\hat{\mathbf{I}},\mathbf{I})\bigr),
\label{eq:rec_loss}
\end{equation}
where $\lambda_{\mathrm{ssim}}$ balances the pixel-wise $\ell_1$ term and the structural-similarity term.
Because haze weakens reliable multi-view geometric cues, we additionally regularize the Gaussian depth with monocular pseudo-depth after per-view scale- and shift-invariant normalization:
\begin{equation}
\mathcal{L}_{\mathrm{depth}}
=
1-\operatorname{Corr}
(\hat{D},D^{m})
+
\lambda_{\mathrm{sm}}
\mathcal{L}_{\mathrm{smooth}},
\label{eq:depth_loss}
\end{equation}
where $\hat{D}$ is the depth rendered by Gaussian alpha compositing, $D^m$ is the monocular pseudo-depth, $\operatorname{Corr}$ denotes Pearson correlation after normalization, $\mathcal{L}_{\mathrm{smooth}}$ is an edge-aware depth regularizer~\cite{yuan2025threeduir}, and $\lambda_{\mathrm{sm}}$ is its weight.

To stabilize early medium decomposition, a dark-channel-derived transmittance prior is used only in Stage I:
\begin{equation}
\mathcal{L}_{\mathrm{dcp}}
=
\|\hat{T}-T_{\mathrm{dcp}}\|_2^2,
\label{eq:dcp_loss}
\end{equation}
where $T_{\mathrm{dcp}}$ is the transmittance prior estimated by the dark-channel prior (DCP)~\cite{he2011darkchannel}.
The Stage-I objective is
\begin{equation}
\mathcal{L}^{(1)}
=
\mathcal{L}_{\mathrm{rec}}
+
\lambda_{\mathrm{depth}}
\mathcal{L}_{\mathrm{depth}}
+
\lambda_{\mathrm{dcp}}
\mathcal{L}_{\mathrm{dcp}},
\label{eq:stage1_loss}
\end{equation}
where $\lambda_{\mathrm{depth}}$ and $\lambda_{\mathrm{dcp}}$ weight the depth and DCP regularization terms, respectively.
MD-TGC is disabled during this stage because the medium transmittance is still evolving and is not yet a reliable optimization signal.

\paragraph{Stage II: supervision-rebalanced refinement.}
After the medium field stabilizes, its parameters are frozen.
The Gaussian scene is refined using
\begin{equation}
\mathcal{L}^{(2)}
=
\mathcal{L}_{\mathrm{rec}}
+
\lambda_{\mathrm{depth}}
\mathcal{L}_{\mathrm{depth}},
\label{eq:stage2_loss}
\end{equation}
while MD-TGC applies Eq.~\eqref{eq:md_tgc} during backpropagation.

The two-stage design separates physical decomposition from optimization rebalancing, enabling clean Gaussian reconstruction under spatially non-uniform haze.

\section{Experiments}
\subsection{Experimental Setup}
\begin{table*}[!t]
\centering
\small
\setlength{\tabcolsep}{3.2pt}
\renewcommand{\arraystretch}{1.05}
\begin{tabular}{c|c|ccc|ccc|ccc}
\hline
\multirow{2}{*}{Type} & \multirow{2}{*}{Method}
& \multicolumn{3}{c|}{RealX3D}
& \multicolumn{3}{c|}{Mip-NeRF 360}
& \multicolumn{3}{c}{Fog-NeRF} \\
\cline{3-11}
& & PSNR$\uparrow$ & SSIM$\uparrow$ & LPIPS$\downarrow$
& PSNR$\uparrow$ & SSIM$\uparrow$ & LPIPS$\downarrow$
& PSNR$\uparrow$ & SSIM$\uparrow$ & LPIPS$\downarrow$ \\
\hline
\multirow{1}{*}{Haze-Agnostic}
& 3DGS
& 11.403 & 0.616 & 0.479
& 14.194 & 0.628 & 0.228
& 13.441 & 0.537 & 0.454 \\
\hline
\multirow{3}{*}{Restoration-Assisted}
& Diff-GS
& 10.590 & 0.617 & 0.510
& 14.556 & 0.622 & 0.293
& 15.515 & 0.605 & 0.386 \\
& IPC-GS
& 12.385 & 0.603 & 0.425
& 19.100 & \underline{0.776} & \underline{0.164}
& 11.672 & 0.544 & \underline{0.351} \\
& BiLaLoRA-GS
& 13.353 & 0.637 & \underline{0.412}
& \underline{20.029} & 0.748 & 0.194
& 14.949 & 0.572 & 0.359 \\
\hline
\multirow{8}{*}{Scattering-Aware}
& Fog-NeRF
& 10.955 & 0.596 & 0.634
& 16.080 & 0.452 & 0.497
& 16.251 & 0.579 & 0.415 \\
& I$^2$-NeRF
& 7.138 & 0.255 & 0.704
& 16.726 & 0.485 & 0.480
& 12.941 & 0.466 & 0.586 \\
& SeaSplat
& 10.662 & 0.543 & 0.568
& 17.839 & 0.684 & 0.285
& 14.740 & 0.564 & 0.497 \\
& WaterSplatting
& 9.719 & 0.462 & 0.677
& 15.593 & 0.553 & 0.373
& 15.589 & 0.559 & 0.502 \\
& Plenodium
& 13.372 & 0.624 & 0.579
& 16.000 & 0.604 & 0.309
& 16.254 & \underline{0.614} & 0.387 \\
& MarineSTD-GS
& 13.663 & 0.621 & 0.581
& 18.109 & 0.687 & 0.249
& \underline{16.644} & 0.611 & 0.396 \\
& 3D-UIR
& \underline{13.988} & \underline{0.647} & 0.468
& 18.593 & 0.724 & 0.239
& 13.888 & 0.583 & 0.426 \\
\cline{2-11}
& Ours
& \textbf{17.338} & \textbf{0.703} & \textbf{0.392}
& \textbf{22.311} & \textbf{0.802} & \textbf{0.162}
& \textbf{17.576} & \textbf{0.679} & \textbf{0.315} \\
\hline
\end{tabular}
\caption{Quantitative comparison on RealX3D, Mip-NeRF 360 foggy scenes, and Fog-NeRF scenes. The suffix ``-GS'' denotes an image-restoration-assisted 3DGS pipeline built on vanilla 3DGS. Best and second-best results are shown in bold and underlined, respectively.}
\label{tab:main_results_complete}
\end{table*}

\paragraph{Datasets.}
We evaluate Tri-DehazeGS on three benchmarks covering real and synthetic haze degradation. \textbf{RealX3D}~\cite{liu2025realx3d} contains real-world multi-view captures with scattering degradation. Following the original protocol, we use eight scenes with hazy views for training and paired clean images for evaluation.
\textbf{Mip-NeRF 360}~\cite{barron2022mipnerf360} is used to evaluate generalization to outdoor and indoor scenes. We select four scenes (counter, kitchen, bicycle, and garden) and synthesize haze using the atmospheric scattering model.
\textbf{Fog-NeRF}~\cite{teigen2023fognerf} provides paired foggy and clean multi-view images with predefined splits. All methods use identical camera poses, image resolutions, and train/test splits.

\begin{figure*}[!t]
\centering
\includegraphics[width=0.98\textwidth]{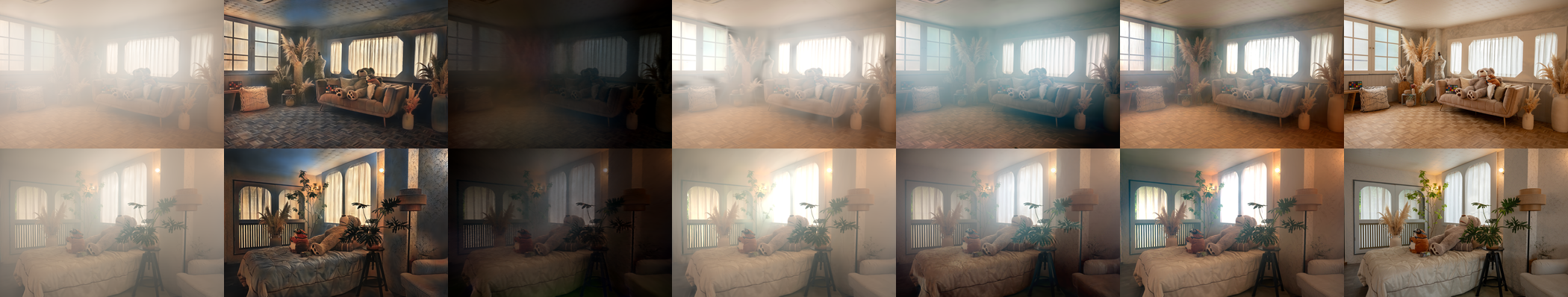}\\[-1pt]
\qualmethodlabels\vspace{-2mm}
\caption{Qualitative comparison of novel view synthesis results on the real RealX3D dataset.}\vspace{-3mm}
\label{fig:real_qualitative_complete}
\end{figure*}

\begin{figure*}[!t]
\centering
\includegraphics[width=0.98\textwidth]{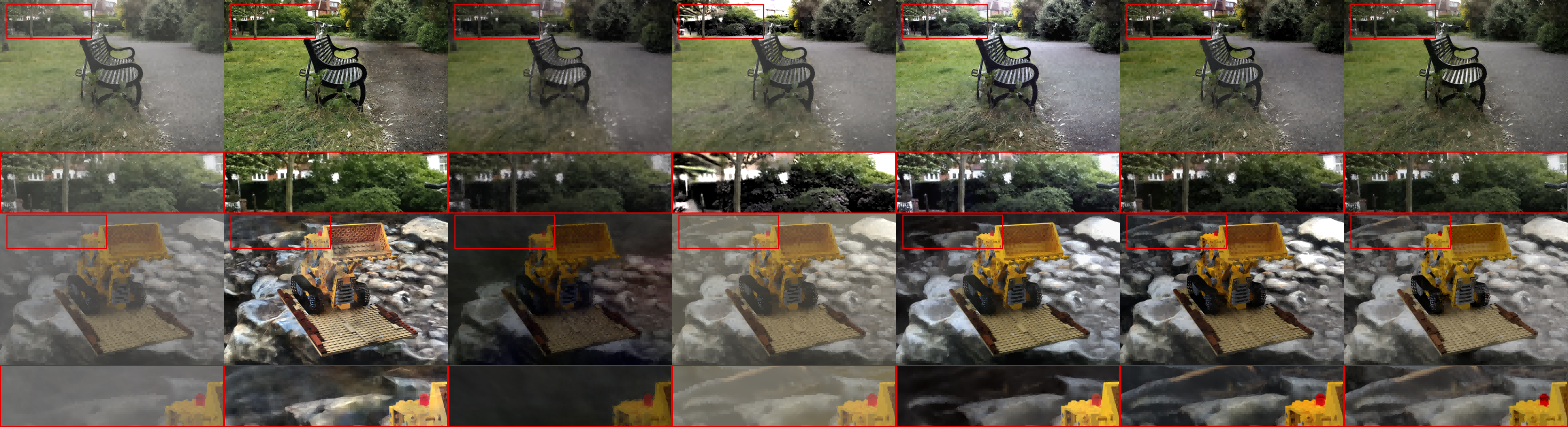}\\[-1pt]
\qualmethodlabels \vspace{-2mm}
\caption{Qualitative comparison of novel view synthesis results on synthetic datasets, including Mip-NeRF 360 foggy scenes and Fog-NeRF scenes.}\vspace{-4mm}
\label{fig:synthetic_qualitative_complete}
\end{figure*}

\paragraph{Baselines.}
We compare against three categories of methods:
(1) \textit{general 3D reconstruction}, i.e., vanilla 3DGS~\cite{kerbl2023gaussian}, 
    to quantify how a haze-agnostic backbone entangles scene and medium;
(2) \textit{restoration-assisted 3DGS pipelines}, including BiLaLoRA~\cite{zhang2026bilalora},
    Diff-Dehazer~\cite{lan2025diffdehazer}, and IPC-Dehaze~\cite{fu2025ipcdehaze},
    to examine whether 2D dehazing can substitute for explicit 3D medium modeling;
(3) \textit{scattering-aware reconstruction}, including Fog-NeRF~\cite{teigen2023fognerf},
    I$^2$-NeRF~\cite{liu2025i2nerf}, SeaSplat~\cite{yang2025seasplat},
    WaterSplatting~\cite{li2025watersplatting}, Plenodium~\cite{wu2025plenodium},
    MarineSTD-GS~\cite{liu2025marinestdgs}, and 3D-UIR~\cite{yuan2025threeduir},
    to isolate the benefit of our view-shared 3D medium field over alternative media formulations.
    
\paragraph{Evaluation Protocol.}
We adopt two complementary protocols. 
\textbf{(i) Restoration} renders novel views from the recovered Gaussian point 
cloud alone and compares them against haze-free ground truth; it applies to 
\emph{all} baselines and serves as the main metric. 
\textbf{(ii) Reconstruction} additionally composites the medium branch onto 
the Gaussian rendering and compares against the original hazy inputs; 
it applies to methods that maintain an explicit 3D scene representation, 
namely vanilla 3DGS and scattering-aware methods (including ours), 
but not to restoration-assisted 3DGS pipelines whose 2D-dehazed inputs 
discard the original medium. Protocol (ii) is reported in the appendix. 

\paragraph{Implementation details.}
We report PSNR, SSIM~\cite{wang2004ssim}, and LPIPS~\cite{zhang2018perceptual}. Higher PSNR/SSIM and lower LPIPS indicate better clean-view reconstruction. All experiments are conducted on NVIDIA RTX 3080 GPUs. Each scene is optimized for 15K iterations. The medium field uses three $64\times64$ tri-planes with 8 channels, two-layer MLP heads, SH order 1 medium light, and 16 ray samples. The medium branch is frozen after 3000 iterations, after which MD-TGC is activated.

\subsection{Main Results}

\paragraph{Quantitative comparison.} Table~\ref{tab:main_results_complete} summarizes comparisons on three benchmarks. Tri-DehazeGS consistently achieves the best reconstruction quality across real and synthetic haze settings. On RealX3D, which contains complex real scattering effects,
Tri-DehazeGS improves over the strongest scattering-aware baseline 3D-UIR by 3.35 dB PSNR and 0.056 SSIM, demonstrating the advantage of a view-shared 3D medium representation for modeling non-uniform haze. Compared with image-restoration-assisted pipelines, Tri-DehazeGS also achieves lower perceptual error, indicating that explicitly modeling the medium is more effective than preprocessing images before reconstruction. On synthetic benchmarks, Tri-DehazeGS improves PSNR by 2.28 dB over
BiLaLoRA-GS on Mip-NeRF 360 foggy scenes and by 0.93 dB over MarineSTD-GS  on Fog-NeRF scenes. These improvements demonstrate that the proposed framework generalizes beyond a specific degradation distribution.

\paragraph{Qualitative comparison.}
Figures~\ref{fig:real_qualitative_complete} and
\ref{fig:synthetic_qualitative_complete} show qualitative comparisons. Existing methods often suffer from residual haze, color distortion, or weak recovery of distant structures.
Medium-aware approaches reduce scattering effects but may still produce inconsistent appearance when the medium is insufficiently decoupled from scene primitives. In contrast, Tri-DehazeGS reconstructs cleaner appearances while preserving stable structures around object boundaries, distant regions, and local textures. These results support the effectiveness of explicitly separating scene radiance from medium effects.

\subsection{Analysis of Scene--Medium Decoupling}

The first analysis investigates whether the improvement originates from explicit medium modeling rather than simply introducing additional parameters.

\begin{table}[t]
\centering
\small
\setlength{\tabcolsep}{1.0pt}
\renewcommand{\arraystretch}{1.08}
\begin{tabular*}{\columnwidth}{@{\extracolsep{\fill}}lcccccc@{}}
\hline
Variant & TP & Decoup. & SH light & PSNR$\uparrow$ & SSIM$\uparrow$ & LPIPS$\downarrow$ \\
\hline
Constant & -- & -- & -- & 13.692 & 0.658 & 0.429 \\
Gaussian-bound & \checkmark & -- & -- & 15.534 & 0.672 & 0.444 \\
Tri-plane RGB & \checkmark & \checkmark & -- & 16.425 & 0.697 & 0.403 \\
Ours & \checkmark & \checkmark & \checkmark & \textbf{17.338} & \textbf{0.703} & \textbf{0.392} \\
\hline
\end{tabular*}
\caption{Ablation of medium representation and scene--medium decoupling.
TP: tri-plane medium field; Decoup: medium field is independent of Gaussian primitives.}
\label{tab:medium_module_ablation_complete}
\end{table}

\begin{figure}[t]
\centering
\includegraphics[width=\columnwidth]{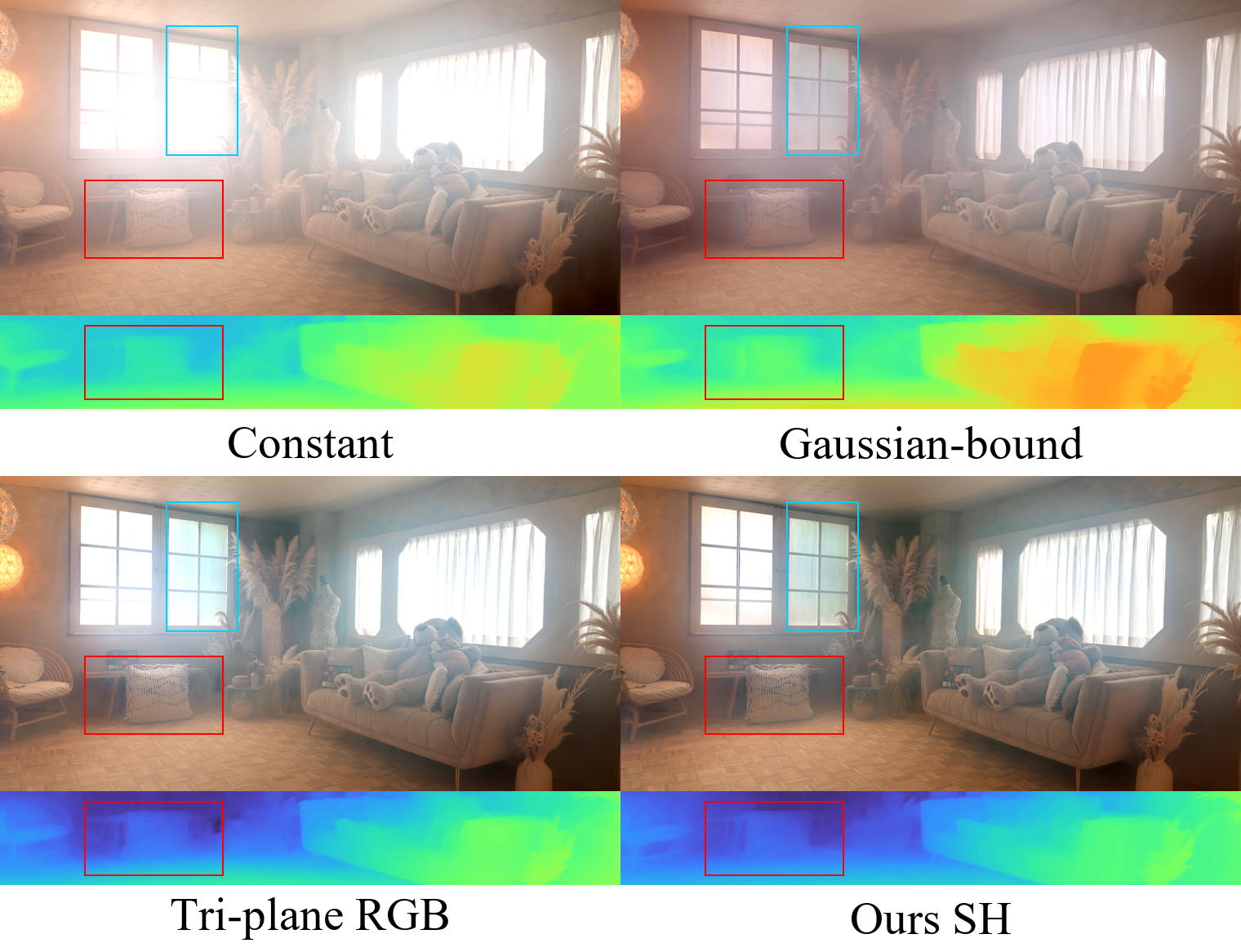}\vspace{-2mm}
\caption{Qualitative ablation of scene--medium coupling and medium-light parameterization on the Akikaze scene. Each variant shows the rendered result and the corresponding transmittance visualization, with highlighted regions indicating local visibility differences.}\vspace{-2mm}
\label{fig:medium_module_ablation_complete}
\end{figure}

Table~\ref{tab:medium_module_ablation_complete} and Figure~\ref{fig:medium_module_ablation_complete} show that the constant medium model performs poorly because it cannot represent spatially varying haze. Coupling the medium field with Gaussian primitives improves over the constant model but remains inferior to explicit decoupling. This confirms that atmospheric effects should not be absorbed into Gaussian appearance. The comparison between Tri-plane RGB and our model further demonstrates that direction-aware medium illumination improves reconstruction under non-uniform lighting.

\begin{figure*}[t]
\centering
{\footnotesize
\makebox[\textwidth][c]{%
\makebox[0.2\textwidth][c]{Coordinate MLP}%
\makebox[0.2\textwidth][c]{Fourier MLP}%
\makebox[0.2\textwidth][c]{Voxel feature}%
\makebox[0.2\textwidth][c]{Ours}%
\makebox[0.2\textwidth][c]{GT}%
}}\\[3pt]
\includegraphics[width=\textwidth]{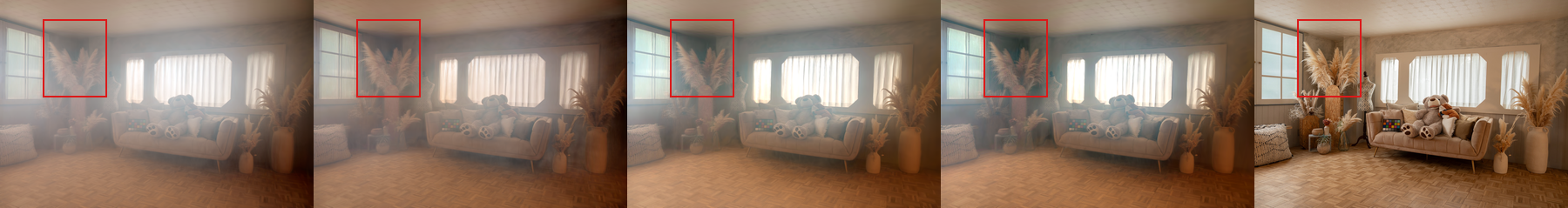}\vspace{-2mm}
\caption{Qualitative comparison of decoupled medium-field parameterizations on the Akikaze scene. All variants use SH1 medium light and differ only in the 3D medium field representation.}\vspace{-3mm}
\label{fig:medium_field_parameterization_main}
\end{figure*}

\paragraph{Medium field parameterization.}
We further compare different 3D medium representations.All variants use the same-depth MLP heads with parameterization-specific inputs, and voxel and tri-plane features share the same resolution. As reported in Table~\ref{tab:medium_field_parameterization_main}, coordinate MLP and Fourier MLP underperform because they struggle to represent spatially varying haze distributions.
Voxel features achieve comparable performance, but the proposed tri-plane representation provides a better balance between compactness and expressiveness. The qualitative comparison in Figure~\ref{fig:medium_field_parameterization_main} shows the same trend. The results indicate that the gain does not come from merely introducing a 3D field, but from constructing an efficient view-shared medium representation.

\begin{table}[t]
\centering
\small
\setlength{\tabcolsep}{6pt}
\renewcommand{\arraystretch}{1.08}
\begin{tabular}{lccc}
\hline
Medium field & PSNR$\uparrow$ & SSIM$\uparrow$ & LPIPS$\downarrow$ \\
\hline
Coordinate MLP & 15.457 & 0.681 & 0.421 \\
Fourier MLP & 14.975 & 0.671 & 0.427 \\
Voxel feature & 17.113 & 0.701 & 0.398 \\
Tri-plane (Ours) & \textbf{17.338} & \textbf{0.703} & \textbf{0.392} \\
\hline
\end{tabular}
\caption{Ablation of decoupled 3D medium-field parameterization. All variants are view-shared, independent of Gaussian primitives, and use SH1 medium light.}
\label{tab:medium_field_parameterization_main}
\end{table}

\subsection{Analysis of Medium-Decoupled TGC}
The second analysis studies whether transmittance can serve as an effective optimization signal. All variants use the same forward haze composition and differ only in the gradient schedule: w/o TGC disables compensation, Full-training TGC (using inverse-transmittance rescaling from the beginning while the medium branch is still trainable), and Ours MD-TGC (activating the compensation only after freezing the medium field).

\begin{figure}[t]
\centering
\includegraphics[width=\columnwidth]{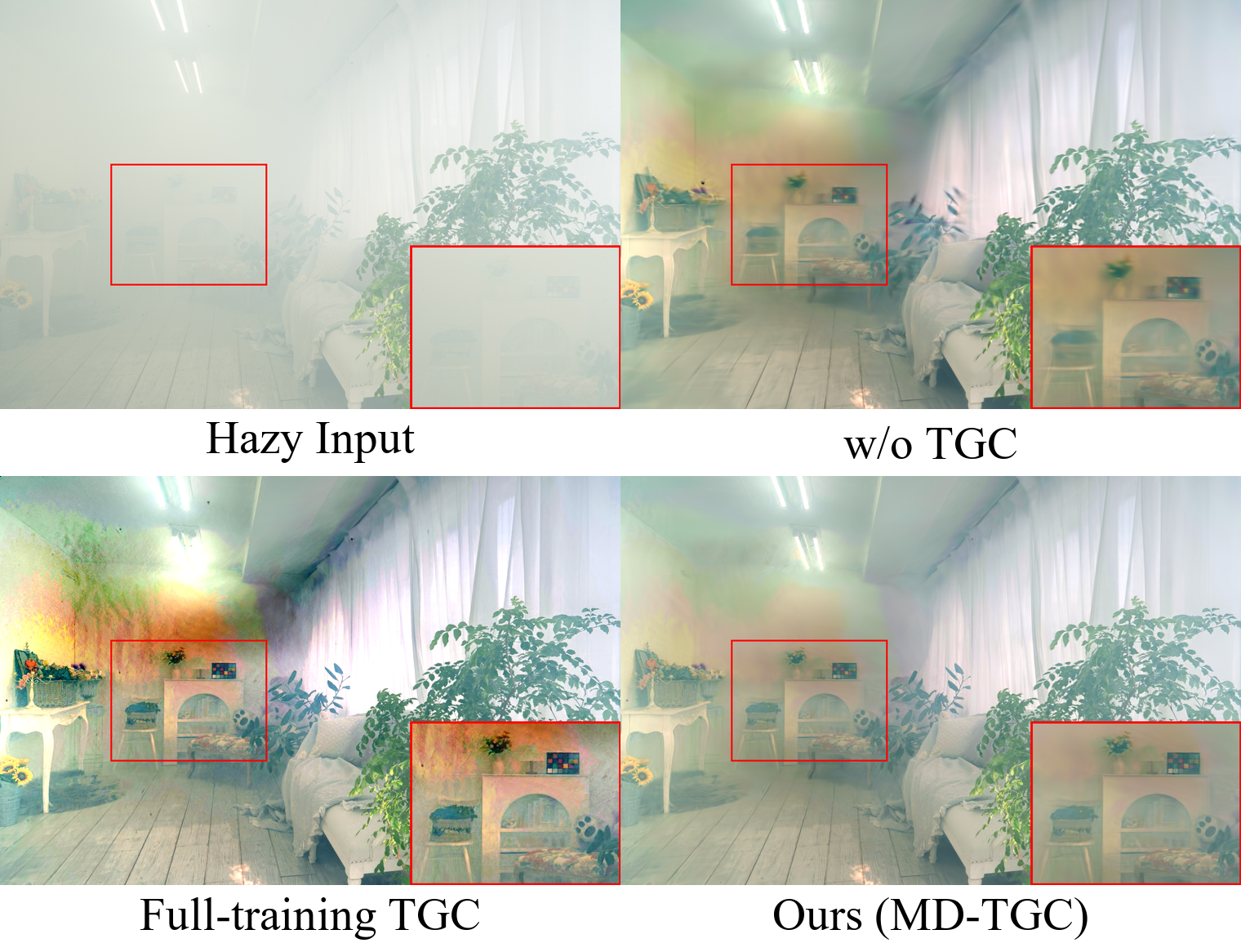}\vspace{-2mm}
\caption{Qualitative ablation of transmittance-gradient-compensation scheduling. The panels show hazy input, w/o TGC, Full-training TGC, and Ours (MD-TGC), with zoomed regions shown in the lower right.}\vspace{-2mm}
\label{fig:md_tgc_ablation_complete}
\end{figure}

\begin{figure}[t]
\centering
\includegraphics[width=\columnwidth]{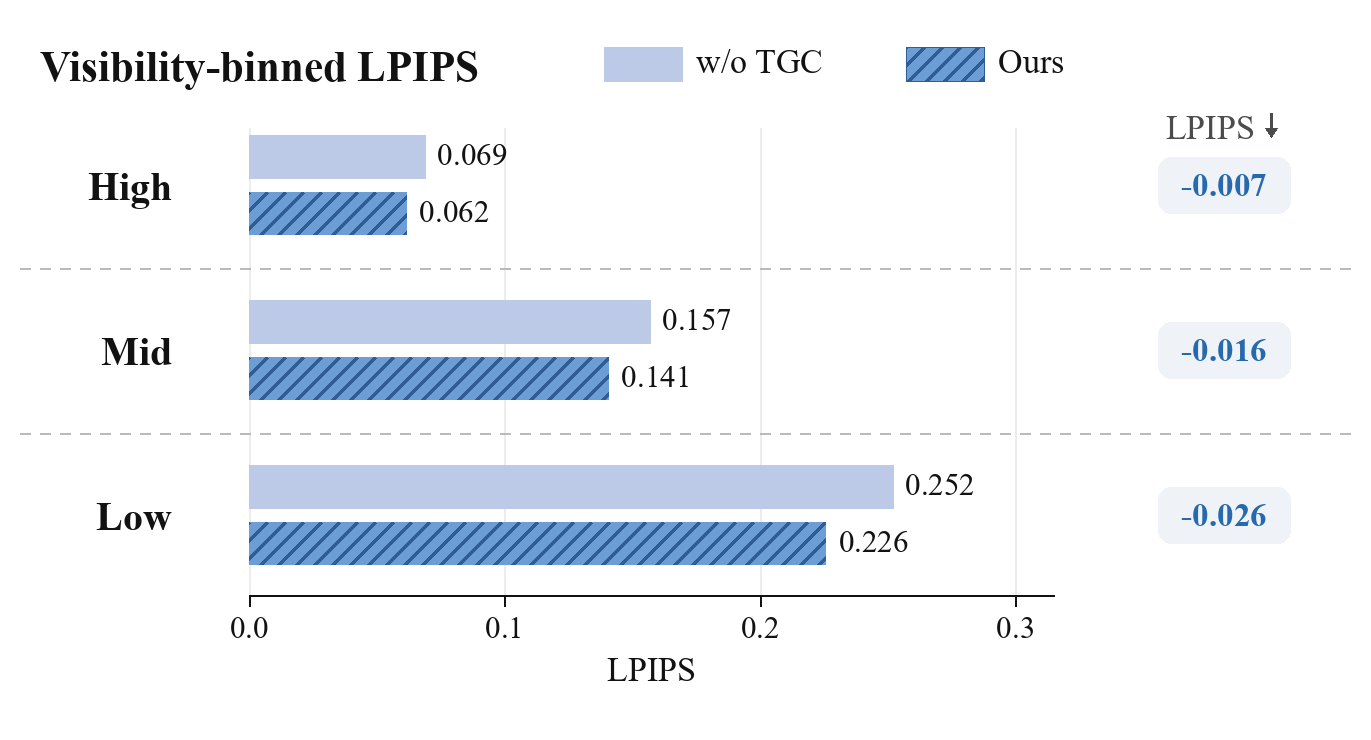}\vspace{-2mm}
\caption{Visibility-binned LPIPS analysis of MD-TGC. High, Mid, and Low denote \(\hat{T}>0.7\), \(0.4<\hat{T}\le0.7\), and \(\hat{T}\le0.4\), respectively. The bars compare w/o TGC and Ours (MD-TGC), and the right-side labels report the decrease in LPIPS from w/o TGC to Ours.}\vspace{-2mm}
\label{fig:md_tgc_visibility_lpips}
\end{figure}

Table~\ref{tab:md_tgc_ablation_complete} compares different compensation strategies. Without TGC, low-transmittance regions receive insufficient gradients. Applying TGC from the beginning improves perceptual quality but decreases PSNR/SSIM because the medium field is still evolving and provides an unstable compensation signal. In contrast, MD-TGC first stabilizes the medium representation and then uses frozen transmittance for gradient compensation, achieving the best overall performance.
Figure~\ref{fig:md_tgc_visibility_lpips} further shows that the improvement in LPIPS becomes larger as visibility decreases, directly validating that MD-TGC mainly benefits regions suffering stronger haze attenuation. Figure~\ref{fig:md_tgc_ablation_complete} shows the same trend: w/o TGC and full-training compensation suffer from unstable colors and artifacts, whereas our schedule recovers cleaner details in low-visibility areas.

\begin{table}[t]
\centering
\small
\setlength{\tabcolsep}{3.5pt}
\renewcommand{\arraystretch}{1.08}
\begin{tabular}{lccc}
\hline
Setting & PSNR$\uparrow$ & SSIM$\uparrow$ & LPIPS$\downarrow$ \\
\hline
w/o TGC & 17.205 & 0.702 & 0.444 \\
Full-training TGC & 16.168 & 0.678 & 0.405 \\
Ours (MD-TGC) & \textbf{17.338} & \textbf{0.703} & \textbf{0.392} \\
\hline
\end{tabular}
\caption{Quantitative ablation of transmittance-gradient-compensation scheduling.}
\label{tab:md_tgc_ablation_complete}\vspace{-2mm}
\end{table}

\subsection{Auxiliary losses}
Finally, we analyze the auxiliary supervision in the two-stage optimization. Table~\ref{tab:loss_ablation_main} shows that removing depth correlation causes the largest drop, indicating that depth priors are important for stabilizing scene--medium decomposition, as weak depth boundaries make complete disentanglement difficult. Removing depth smoothness or DCP causes smaller drops, suggesting that they mainly act as auxiliary stabilizers. Nevertheless, the preceding ablations show that depth priors alone are insufficient; explicit scene--medium decoupling and MD-TGC remain necessary for high-quality clean reconstruction.
\begin{table}[t]
\centering
\small
\setlength{\tabcolsep}{6pt}
\renewcommand{\arraystretch}{1.08}
\begin{tabular}{lccc}
\hline
Setting & PSNR$\uparrow$ & SSIM$\uparrow$ & LPIPS$\downarrow$ \\
\hline
w/o Depth Corr. & 12.767 & 0.644 & 0.451 \\
w/o Depth Smooth & 17.023 & 0.702 & 0.405 \\
w/o DCP & 16.894 & 0.702 & 0.397 \\
Ours & \textbf{17.338} & \textbf{0.703} & \textbf{0.392} \\
\hline
\end{tabular}
\caption{Ablation on auxiliary training losses.}\vspace{-2mm}
\label{tab:loss_ablation_main}
\end{table}

\section{Limitations}
\begin{figure}[t]
\centering
\includegraphics[width=\columnwidth]{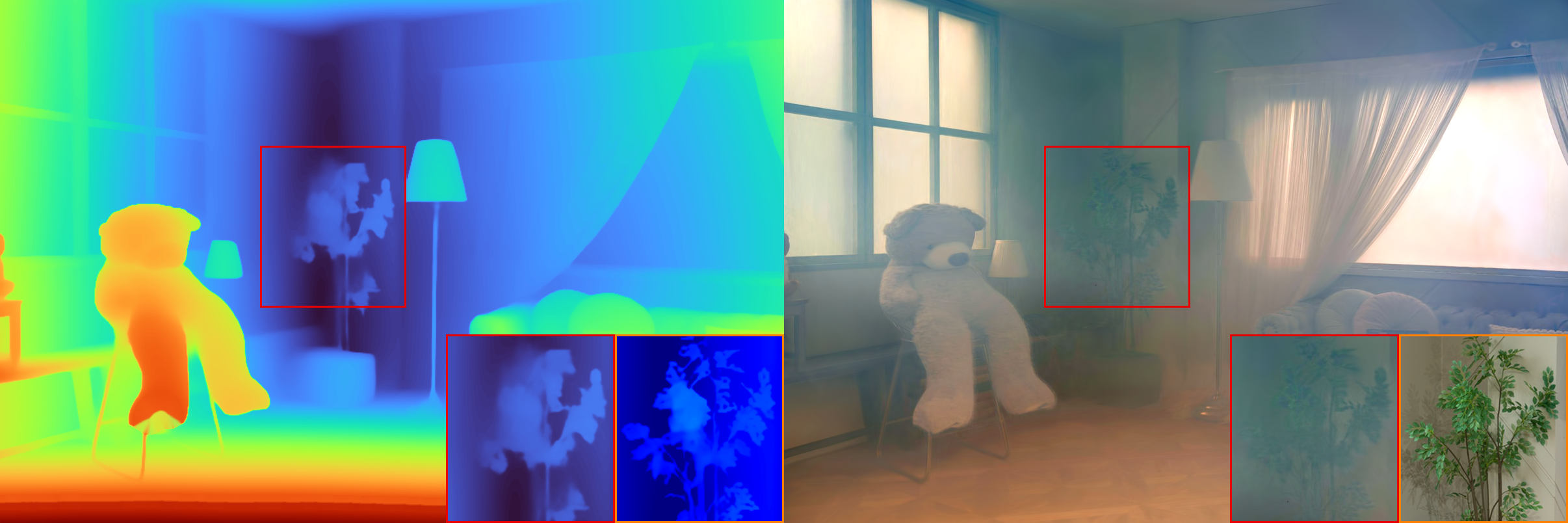}\\[3pt]
{\footnotesize
\makebox[\columnwidth][c]{%
\makebox[0.5\columnwidth][c]{Pseudo depth}%
\makebox[0.5\columnwidth][c]{Clean}%
}}
\vspace{-2mm}
\caption{Failure case due to the inaccurate depth prior. }
\vspace{-2mm}
\label{fig:limit}
\end{figure}

Tri-DehazeGS leverages an estimated depth prior to regularize the scene--medium decomposition and stabilize the optimization process. Consequently, the final restoration quality is influenced by the accuracy of the estimated depth. As illustrated in  Figure~\ref{fig:limit}, inaccurate or over-smoothed depth estimates, particularly around thin structures and low-texture regions, may result in blurred object boundaries and the loss of fine details in the recovered clean scene. This limitation indicates that incorporating more reliable depth priors or structure-aware constraints could further enhance reconstruction quality in such ambiguous regions.

\section{Conclusion}
In this work, we study clean 3D Gaussian reconstruction from hazy multi-view observations and identify two key challenges: scene--medium entanglement in representation learning and transmittance-induced supervision attenuation during optimization. To address these issues, we propose Tri-DehazeGS, which explicitly separates scene radiance from participating-medium effects through a view-shared three-dimensional medium representation and introduces Medium-Decoupled Transmittance Gradient Compensation (MD-TGC) to restore optimization signals suppressed by haze. Extensive experiments on real and synthetic haze benchmarks demonstrate that Tri-DehazeGS consistently improves clean novel-view synthesis quality over existing scattering-aware and restoration-assisted baselines. Further analyses verify that the performance gains arise from both improved scene--medium decomposition and transmittance-aware optimization rebalancing. More broadly, this work suggests that physically degraded neural reconstruction should be viewed as a joint problem of representation disentanglement and optimization rebalancing, rather than solely an image formation modeling problem.

\bibliography{references}

\end{document}